\documentclass[10pt,twocolumn,letterpaper]{article}

\usepackage[algorithms]{wacv}      %

\usepackage{amsmath}
\usepackage{amssymb}
\usepackage{amsthm}
\usepackage{booktabs}
\usepackage{multirow}
\usepackage{xspace}

\AtEndPreamble{%
  \crefname{appendix}{Sec.}{Secs.}%
  \Crefname{appendix}{Section}{Sections}%
}

\newtheorem{theorem}{Theorem}
\newtheorem{definition}{Definition}

\newcommand{\R}{\mathbb{R}}
\newcommand{\SOtwo}{\mathrm{SO}(2)}
\newcommand{\E}{\mathbb{E}}
\newcommand{\expmop}{\mathrm{expm}}
\newcommand{\rhotil}{\tilde{\rho}}

\DeclareMathOperator*{\argmin}{arg\,min}

\newcommand{\twofigpage}[6]{%
  \twocolumn[{%
    \centering
    #1\par
    \vspace{0.4em}%
    \refstepcounter{figure}\label{#2}%
    \begin{minipage}{0.95\textwidth}%
      \small\textbf{Figure~\thefigure.}\enspace #3%
    \end{minipage}\par
    \vspace{1.2em}%
    #4\par
    \vspace{0.4em}%
    \refstepcounter{figure}\label{#5}%
    \begin{minipage}{0.95\textwidth}%
      \small\textbf{Figure~\thefigure.}\enspace #6%
    \end{minipage}\par
    \vspace{1.0em}%
  }]%
}

\definecolor{wacvblue}{rgb}{0.21,0.49,0.74}
\usepackage[pagebackref,breaklinks,colorlinks,allcolors=wacvblue]{hyperref}

\def\wacvPaperID{406}
\def\confName{WACV}
\def\confYear{2027}

\title{Pair-Relative Partial Equivariance for Rotation-Robust\\ Keypoint Descriptor Matching}

\author{Jongmin Lee\\
Chung-Ang University\\
{\tt\small jmlee@cau.ac.kr}
}

\begin{document}
\maketitle
\begin{abstract}
Learned keypoint descriptors trained on upright imagery often fail under large
in-plane rotation. Steerers~\cite{bokman2024steerers} address this by applying a
single exact $\SOtwo$ representation to the descriptor space, but this global
group constraint can be too rigid for keypoint descriptors, whose angular
responses need not follow a single group representation. We propose pair-relative
partial equivariance, which measures equivariance under the empirical
distribution of relative rotations in image pairs, and PaSteer, a relaxed
descriptor-steering framework that allows the steering map to leave the exact
representation manifold while retaining the lightweight, backbone-free design of
steering. PaSteer instantiates this relaxation with two variants: an
identity-preserving residual steerer (PaSteer-soft) and a free-form Fourier
steerer (PaSteer-free). We show that the per-angle optimal linear steerer for a
fixed non-equivariant descriptor generally does not obey the group composition
law, explaining why exact representations can be suboptimal in the
frozen-descriptor regime. On Roto-360, MegaDepth-1500, AIMS, and IMC-PT, PaSteer
consistently improves rotation-robust matching over an exact steerer on frozen
DeDoDe descriptors, especially under large relative rotations, with minimal
upright degradation.

\end{abstract}

\section{Introduction}
\label{sec:intro}

Local feature matching is a backbone of structure-from-motion, visual
localization, and 3D reconstruction. Modern learned descriptors such as
SuperPoint \cite{detone2018superpoint}, DISK \cite{tyszkiewicz2020disk}, XFeat
\cite{potje2024xfeat}, and DeDoDe \cite{edstedt2024dedode} are trained almost
exclusively on upright internet and SfM imagery. Unlike SIFT
\cite{lowe2004sift}, they therefore degrade sharply once the relative in-plane
rotation between two images leaves a narrow band. This matters in practice:
astronaut photography from the ISS \cite{stoken2023aims}, endoscopy, underwater
imagery, and aerial imagery routinely contain large or arbitrary roll.

\begin{figure}[t]
\centering
\includegraphics[width=\linewidth]{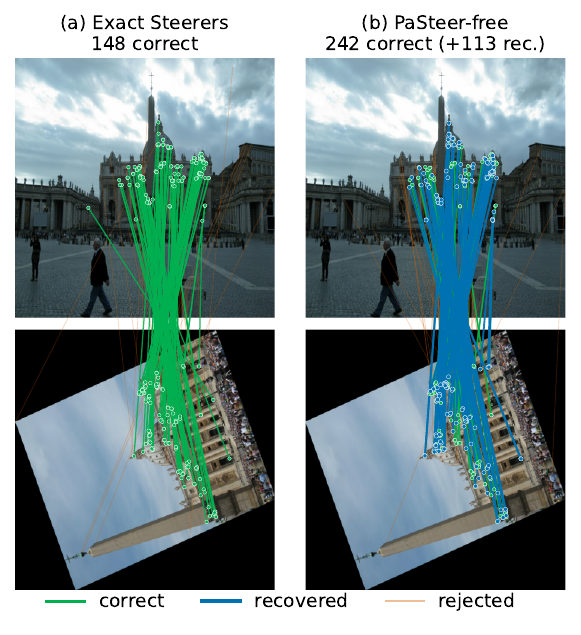}
\vspace{-0.9cm}
\caption{\textbf{PaSteer-free recovers correspondences that exact steering misses.}
A MegaDepth-1500 \cite{li2018megadepth} pair under a large
$112^\circ$ in-plane rotation. Both panels share the \emph{same} frozen DeDoDe-B
\cite{edstedt2024dedode} descriptor; only the steering changes. Exact Steerers
\cite{bokman2024steerers} finds 148 correct matches (left), PaSteer-free 242 (right).
Green: \textit{correct}; blue: \emph{recovered} ($+113$ correct matches); %
orange: \textit{rejected}.}
\label{fig:teaser}
\end{figure}

Existing remedies either build rotation handling into the descriptor or add
steering on top of a released descriptor. Rotation-invariant or equivariant
descriptors improve robustness through group pooling \cite{liu2019gift} or
rotation-equivariant backbones with group alignment
\cite{lee2022rekd,lee2023relf,santellani2023strek}, but they require modifying
or retraining the representation and may discard orientation as a useful
matching cue. Steerers \cite{bokman2024steerers} take a lighter route: they keep
the descriptor backbone fixed and encode rotation as a linear map
$\rho(\theta)$ on the description space. However, this map is constrained to be
an \emph{exact} $\SOtwo$ representation,
$\rho(\theta+\phi)=\rho(\theta)\rho(\phi)$, forcing one global homomorphism to
explain the descriptor response over the entire group. For an already-trained
descriptor that was not optimized to be equivariant, this exact representation
constraint can be overly rigid.  Throughout, by a \emph{frozen released} descriptor we mean an off-the-shelf descriptor taken from publicly released weights and used as is, without any retraining.

The key observation of this paper is that rotation robustness in matching is
pair-relative, not group-uniform. Matching depends on the relative rotation
$\Delta\theta$ between two images, and the distribution of $\Delta\theta$ varies
sharply by domain. Measured from ground-truth poses, MegaDepth-1500~\cite{li2018megadepth} pairs are
mostly near-upright, whereas astronaut imagery~\cite{stoken2023aims} is close to uniform over
$\SOtwo$ (\cref{sec:experiments}). Thus, a single fixed degree of equivariance
cannot be ideal for both. This is precisely the setting addressed by
\emph{partial equivariance}, where equivariance is not all-or-nothing but a
data-dependent degree
\cite{romero2022partial,benton2020augerino,finzi2021rpp,wang2022approxequiv,veefkind2024probdegree}.
To our knowledge, this view has not been applied to descriptor matching.

We connect partial equivariance to rotation-robust descriptor matching through
\emph{PaSteer}, a relaxed steering framework that keeps the lightweight,
backbone-free design of descriptor steering while loosening the exact $\SOtwo$
representation constraint. PaSteer allows the steering map to leave the
representation manifold, rather than forcing one homomorphism to fit all
angles. We instantiate this idea in two operating points:
\emph{PaSteer-soft}, a conservative identity-preserving relaxation anchored to
the exact generator, and \emph{PaSteer-free}, a free-form Fourier steerer that
directly models the angle-dependent map. Both variants change only the steering
map; the detector, frozen descriptor, keypoints, probe grid, matcher, and
geometric estimator are kept fixed in our controlled comparisons.
Across Roto-360~\cite{lee2022rekd}, MegaDepth-1500~\cite{li2018megadepth}, AIMS~\cite{stoken2023aims}, and IMC-PT~\cite{jin2021image}, this relaxation improves
frozen-descriptor matching most under large relative rotations.

Our contributions are four-fold:
\begin{itemize}\itemsep2pt
\item We formulate \emph{pair-relative partial equivariance} for descriptor
matching, measuring equivariance against the empirical distribution
$q(\Delta\theta)$ of pairwise relative rotations rather than against the
uniform group.
\item We introduce \textbf{PaSteer}, a relaxed steering framework that
preserves the deployability of descriptor steering while loosening the exact
$\SOtwo$ representation constraint.
\item We show that, for a frozen non-equivariant descriptor, the per-angle
optimal linear steerer is generally non-homomorphic, so an exact $\SOtwo$
representation is intrinsically restrictive.
\item We clarify the scope of relaxation: it is most beneficial for frozen
released descriptors and becomes comparable to exact steering under full
joint descriptor--steerer training.
\end{itemize}

\section{Related Work}
\label{sec:related}

\paragraph{Local feature matching.}
Establishing correspondences between images underlies structure-from-motion,
visual localization, and 3D reconstruction. Sparse pipelines detect and describe
keypoints \cite{lowe2004sift,detone2018superpoint,tyszkiewicz2020disk,potje2024xfeat},
and DeDoDe \cite{edstedt2024dedode} in particular decouples detection from
description and is the descriptor we build on. Learned matchers then resolve the
assignment, either over sparse keypoints
\cite{sarlin2020superglue,lindenberger2023lightglue} or densely without an
explicit detector \cite{sun2021loftr,edstedt2024roma}. These descriptors and
matchers are trained almost exclusively on upright internet and SfM imagery, so
they degrade once the relative rotation of a pair leaves a narrow band; dense
matchers such as RoMa \cite{edstedt2024roma} are robust to viewpoint but remain
weak under large in-plane rotation. We propose neither a new detector nor a new
matcher: our intervention is confined to the descriptor steering stage, leaving
the detector and matcher fixed so that steering mechanisms can be compared on a
single frozen descriptor under controlled conditions.

\paragraph{Rotation-equivariant features and descriptors.}
Group-equivariant CNNs extend weight sharing from translations to larger symmetry
groups \cite{cohen2016gcnn,worrall2017hnet}, and steerable formulations express
the rotation action analytically on feature fields
\cite{cohen2017steerable,weiler2019e2cnn}. For matching, one line builds rotation
handling directly into the descriptor representation. GIFT \cite{liu2019gift}
pools a descriptor over a transformation orbit to obtain invariance,
Self-Sca-Ori \cite{lee2021selfsca} estimates a canonical keypoint scale and orientation, while
REKD \cite{lee2022rekd}, RELF \cite{lee2023relf}, and S-TREK
\cite{santellani2023strek} learn rotation-equivariant detectors and descriptors
and align the group dimension to a canonical pose; RELF reports strong results on
the Roto-360 benchmark introduced with REKD \cite{lee2022rekd}. The same
group-equivariant machinery underlies dense correspondence \cite{min2020dhpf} and
frequency-domain pose regression \cite{lee2024wignerd}. SE2-LoFTR
\cite{bokman2022se2loftr} instead places an $\mathrm{SE}(2)$-equivariant backbone
inside a detector-free matcher. Building the symmetry into the representation is
robust, but it carries a cost: invariance spends discriminative capacity, since
orientation is itself a matching cue, and an equivariant backbone must be trained
from scratch rather than reused from a released descriptor. These methods also
impose equivariance over the entire group.

\paragraph{Steerers and orientation handling.}
Steerers \cite{bokman2024steerers} are the closest baseline to our setting: they
keep the backbone untouched and encode a rotation as a single linear map on the
description space of a frozen or jointly trained DeDoDe descriptor, requiring
$\rho$ to be an exact representation of $\SOtwo$; they report state of the art on
Roto-360 and AIMS~\cite{stoken2023aims}. Requiring an exact representation ties a global $\SOtwo$ action
to the whole descriptor space, so one linear map must account for every rotation
consistently. Affine Steerers \cite{bokman2024affinesteerers} generalize the
construction to $\mathrm{GL}(2)$. A follow-up line asks whether invariance should
live in the descriptor or the matcher \cite{bokman2026orientation,nordstrom2026loma}
and shows that data scaling shrinks the upright-rotated gap. These works steer with
an exact representation, train with multiples of $90^\circ$, and do not train
descriptors with non-trivial steerers. We ask instead whether an exact
representation is the right constraint for a frozen released descriptor, and relax
it to a continuous, pair-relative degree of equivariance.

\paragraph{Partial, relaxed, and learned equivariance.}
A growing body of work treats equivariance as a learnable degree rather than a
hard constraint. Augerino \cite{benton2020augerino} learns an augmentation
distribution; Residual Pathway Priors \cite{finzi2021rpp} add a soft
non-equivariant residual path; approximately equivariant networks
\cite{wang2022approxequiv} mix equivariant and free kernels; and Partial G-CNNs
\cite{romero2022partial} learn a distribution over a group subset so a layer is
equivariant only on $S\subseteq G$. The probabilistic-degree approach
\cite{veefkind2024probdegree} learns per-layer equivariance for steerable CNNs via
a Fourier parameterization. The Lie derivative \cite{gruver2023liederivative}
quantifies learned equivariance, and approximation-generalization trade-offs under
approximate group equivariance have been analyzed theoretically
\cite{petrache2023approxgen,kondor2018generalization}. All of these methods relax
the degree of equivariance inside a network backbone, and none has been applied to
image matching or to descriptor steering. PaSteer adapts the partial-equivariance
idea to the steering map rather than the backbone, making the degree of
equivariance data-determined and tying it to the empirically measured
\emph{relative} rotation distribution of image pairs.

\section{Method}
\label{sec:method}

\begin{figure}[t]
  \vspace{-0.2cm}
  \centering
  \includegraphics[width=\linewidth]{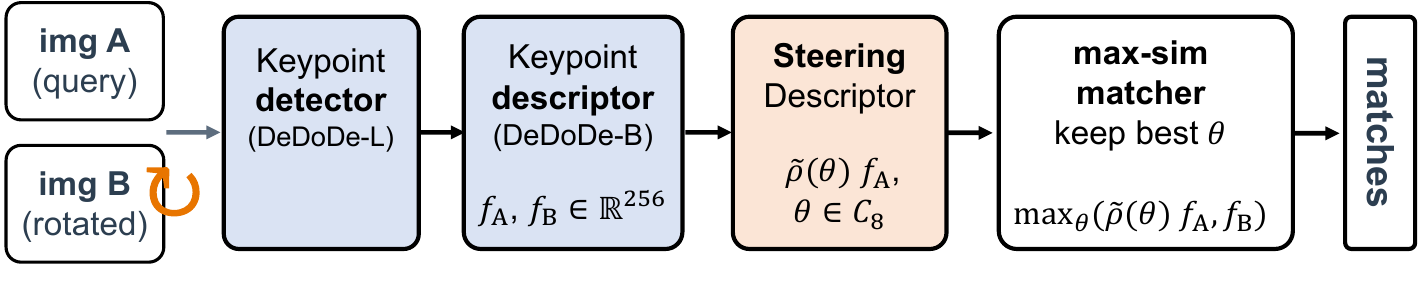}
  \vspace{-0.8cm}
  \caption{\textbf{PaSteer matching pipeline.} An $\SOtwo$ keypoint detector (DeDoDe-L~\cite{edstedt2024dedode})
  and a keypoint descriptor (DeDoDe-B) feed a steered max-similarity matcher:
  the query descriptors are steered by $\rhotil(\theta)$ over the $C_8$ probe
  grid under partial equivariance, and the best-scoring angle is kept per keypoint. }
  \label{fig:pipeline}
\end{figure}

\Cref{fig:pipeline} overviews the six-stage pipeline. \textbf{(1)} A query $I_1$ and
a target $I_2$, which may differ by an arbitrary in-plane rotation, enter the system.
\textbf{(2)} A rotation-equivariant $\SOtwo$ detector (DeDoDe-L~\cite{edstedt2024dedode}) produces keypoints in
both images. \textbf{(3)} A \emph{frozen} DeDoDe-B descriptor $f$ describes each
keypoint, giving $f_A,f_B\in\R^{256}$. \textbf{(4)} Before matching, a steering map
evaluated over a $C_8$ probe grid steers the query descriptors into a bank of candidate
frames. \textbf{(5)} A maximum-similarity matcher scores each steered query against the
target, keeps the best-scoring angle per keypoint, and applies a
mutual-nearest-neighbour selection. \textbf{(6)} The surviving correspondences feed the
usual geometric estimator (essential matrix or homography) for pose or registration.

The rest of this section develops PaSteer. \Cref{sec:pairrel} formalizes
\emph{pair-relative partial equivariance}, the degree of equivariance measured
against the rotation a domain actually exhibits, and gives the intuition for why
relaxing helps; \cref{sec:relaxed} introduces the relaxed steering family and its two
instances, \cref{sec:steering} how each steerer fills the steering slot,
\cref{sec:matcher} the matcher that turns steered descriptors into correspondences, and
\cref{sec:training} explains training objectives.

\subsection{Pair-relative partial equivariance}
\label{sec:pairrel}

Let $f:X\to\R^D$ be a keypoint descriptor on an image domain $X$, and let
$R_\theta:X\to X$ be the in-plane rotation action for $\theta\in\SOtwo\cong
[0,2\pi)$. A \emph{steerer}~\cite{bokman2024steerers} is a map $\rho:\SOtwo\to\R^{D\times D}$ intended to
satisfy $f(R_\theta I)\approx\rho(\theta)f(I)$, so that a descriptor extracted
from an upright image can be transported into the frame of a rotated one. In
exact steering \cite{bokman2024steerers}, $\rho$ is constrained to be a group
representation,
\begin{equation}
  \rho(\theta+\phi)=\rho(\theta)\rho(\phi),\qquad \rho(0)=I,
  \label{eq:rep}
\end{equation}
which, for $\SOtwo$, is parameterized by a single generator $D$ via
$\rho(\theta)=\expmop(\theta D)$ with $\expmop(2\pi D)=I$.

The key observation is that matching performance depends on the \emph{relative}
rotation $\Delta\theta=\theta_2-\theta_1$ of a pair $(I_1,I_2)$, not on absolute
orientation. We therefore measure equivariance against the empirical
relative-rotation distribution of the data~\cite{gruver2023liederivative}, following the
partial-equivariance view~\cite{romero2022partial,benton2020augerino,finzi2021rpp,wang2022approxequiv,veefkind2024probdegree,petrache2023approxgen}.

\begin{figure}[t]
  \centering
  \scalebox{0.9}{
  \includegraphics[width=\linewidth]{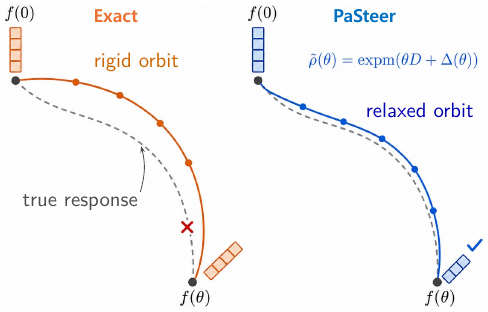}
  }
  \vspace{-0.3cm}
  \caption{\textbf{Exact representation vs.\ relaxed steering.} The gray dashed curve denotes the true angular response of the keypoint descriptor. The orange solid curve shows the rigid exact-representation orbit, which misses this response, while the blue solid curve shows the relaxed PaSteer-soft orbit,
$\rhotil(\theta)=\expmop(\theta D+\Delta(\theta))$, which follows the true response more closely.}
  \label{fig:relax}
\end{figure}

\begin{definition}[Pair-relative partial equivariance]
\label{def:pairrel}
Let $q(\Delta\theta)$ be the distribution of the relative rotation of image pairs
in a domain. A descriptor $f$ is $\varepsilon$-partially equivariant with respect
to $q$ and a steerer $\rho$ if
\begin{equation}
  \E_{\Delta\theta\sim q}\,\E_I\;\big\|\,f(R_{\Delta\theta} I)-\rho(\Delta\theta)\,f(I)\,\big\|^2 \le \varepsilon.
  \label{eq:pairrel}
\end{equation}
\end{definition}

The exact-steering setting is the special case $q=\mathrm{Uniform}(\SOtwo)$ with
$\rho$ constrained as in \cref{eq:rep}. We make two changes: $q$ is estimated
from data, concentrated near $0^\circ$ for SfM imagery and near-uniform for
astronaut imagery, and $\rho$ is allowed to leave the representation manifold. We
refer to the constrained case as \emph{exact} and the unconstrained case as
\emph{relaxed}.

\begin{figure}[t]
  \centering
  \includegraphics[width=0.75\linewidth]{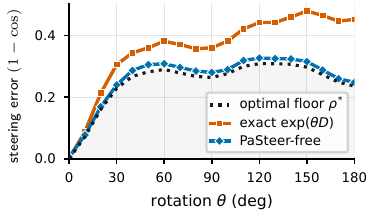}
  \vspace{-0.4cm}
  \caption{\textbf{The relaxed map approaches the optimal floor.} Steering error is $1-\cos$ of the cosine similarity between a steered upright descriptor $\rho(\theta)f(I)$ and the truly rotated one $f(R_\theta I)$ (lower is better, $0$ is exact recovery). On fixed DeDoDe-B~\cite{edstedt2024dedode} descriptor,
  PaSteer tracks the optimal per-angle floor, while the \textit{exact} representation~\cite{bokman2024steerers} falls short and its error grows with rotation angle.}
  \label{fig:floor}
\end{figure}

\noindent\textbf{Why relaxing helps.} \Cref{fig:relax} shows the intuition. The
\textit{frozen} keypoint descriptor was never trained to rotate as a group, so its true
angular response is not a representation, and the \emph{best} linear steerer at each
angle is not one either. An exact representation is locked to the representation
manifold and cannot reach this per-angle optimum, whereas a relaxed steerer can.
\Cref{fig:floor} visualizes the steering error of the exact map growing with angle while the
relaxed map hugs the optimal per-angle floor. Supp.\ \cref{sec:supp-theory} makes this precise
with a closed form for the optimal per-angle steerer.

\subsection{Relaxed steering family}
\label{sec:relaxed}

\textbf{PaSteer} is a relaxed steering family that lets the steering map leave the
exact representation manifold, placing a single knob between a rigid exact steerer and a
fully free-form one and turning it only as far as the data needs. It has two instances,
named consistently throughout the paper.

\noindent\textbf{PaSteer-soft} is a residual-exponential relaxation \emph{anchored} to
the exact generator: it keeps the generator-exponential form and adds only a small
low-rank Fourier residual that vanishes at $\theta=0$, in the spirit of
soft/approximate-equivariance residuals~\cite{finzi2021rpp,wang2022approxequiv}. This
preserves the upright identity exactly, while relaxing the exact representation
constraint with a small number of additional parameters. It is the conservative,
capacity-light, identity-preserving variant.

\noindent\textbf{PaSteer-free} is a \emph{free-form} Fourier relaxation: since the
per-angle optimal steerer is in general not a representation (supp.\ \cref{sec:supp-theory}),
it drops the generator-exponential form and parameterizes the angle-dependent map
directly in a Fourier basis over the probe grid~\cite{veefkind2024probdegree,romero2022partial},
giving up the orthogonality and homomorphism constraints to approach that optimum. It is
the main high-capacity instance in our full-scale experiments; its upright identity is
approximate rather than exact, and its larger parameterization is a training and storage
cost.

With the exact representation as the rigid endpoint, the three steerers used
throughout the paper are:
\begin{itemize}\itemsep0pt
  \item \textbf{Exact}: a single $\SOtwo$ representation.
  \item \textbf{PaSteer-soft}: a representation-\emph{anchored}, bounded residual-exponential relaxation.
  \item \textbf{PaSteer-free}: a \emph{free-form} Fourier relaxation that drops the representation form.
\end{itemize}
``Exact'' denotes the Steerers \cite{bokman2024steerers} representation retrained at
our budget, used as the baseline throughout.

\subsection{Steering stage}
\label{sec:steering}

All compared methods share the same detector, frozen descriptor, keypoints, $C_8$
probe grid, matcher, and geometric estimator; the only component that changes is the
steering map $\rhotil$. \textbf{Exact} fills this slot with the representation
$\rho(\theta)=\expmop(\theta D)$. \textbf{PaSteer-soft} adds a bounded low-rank Fourier
residual to the exact generator core,
\begin{equation}
  \rhotil(\theta)=\expmop\!\big(\theta D+\Delta(\theta)\big),
  \label{eq:relaxed}
\end{equation}
\begin{equation}
\begin{aligned}
  \Delta(\theta)
  &= \sum_{k=1}^{K}\big[\,S_k\sin k\theta+C_k(\cos k\theta-1)\,\big], \\
  S_k &= U_kV_k^\top,\qquad
  C_k = P_kQ_k^\top \quad \text{of rank } r .
\end{aligned}
\label{eq:residual}
\end{equation}
with $\Delta(0)=0$, so $\rhotil(0)=I$ exactly and upright pairs incur no steering
error. \textbf{PaSteer-free} drops the representation form and parameterizes the
per-angle map directly,
\begin{equation}
  \rhotil(\theta)=A_0+\sum_{k=1}^{K}\big[\,A_k\cos k\theta+B_k\sin k\theta\,\big],\quad A_0\approx I,
  \label{eq:freefourier}
\end{equation}
an unconstrained linear map per angle. Each steerer then applies its chosen map over
the $C_8$ probe grid $\Theta=\{0,45^\circ,\dots,315^\circ\}$, sending a query
descriptor $f_A$ to a steered bank $g_A(\theta)=\rhotil(\theta)f_A$.

\noindent\textbf{Deployability.} All three deployed steerers share the same test-time
structure: their $C_8$ steering maps are precomputed before inference, and during
matching each query descriptor is transformed by these fixed maps and scored by the
same maximum-similarity matcher. PaSteer-free therefore has identical inference cost to
exact steering; its larger parameterization affects only training and storage, not the
per-pair inference. We separately test whether the gain comes from extra parameters in \cref{sec:mechanisms}.

\subsection{Matching}
\label{sec:matcher}
We score steered descriptors over candidate angles and produce final
correspondences. For each query keypoint, the matcher selects the candidate
angle whose steered descriptor best matches the target, following the steered
maximum-similarity matching of Steerers~\cite{bokman2024steerers}.
With $\ell_2$-normalized descriptors, the similarity between a query keypoint
$a$ and a target keypoint $b$ is
\begin{equation}
S_{ab}=\max_{\theta\in\Theta}
\big\langle \widehat{\rhotil(\theta)f_A^{a}},\widehat{f_B^{b}}\big\rangle .
\label{eq:matching}
\end{equation}
An upright pair selects an angle near $\theta=0$, whereas a rotated pair selects
the angle that best explains its relative rotation, without requiring any
rotation prior. A mutual-nearest-neighbour dual-softmax
selection~\cite{sun2021loftr} on $S$ yields the final correspondences for the
geometric estimator. Since Exact, PaSteer-soft, and PaSteer-free differ only in
$\rhotil$, this comparison isolates the effect of relaxing the steering map.

\subsection{Training}
\label{sec:training}

\noindent\textbf{Frozen-descriptor objective.} With $f$ fixed, we train only the
steering map $\rhotil$ by minimizing the steering error of \cref{def:pairrel}.
\Cref{eq:pairrel} uses a squared $\ell_2$ distance for clarity; we minimize its
empirical cosine form, consistent with the matcher of \cref{sec:matcher},
$\mathcal{L}=\E_{\Delta\theta\sim q}\,\E_I[\,1-\cos(f(R_{\Delta\theta}I),\rhotil(\Delta\theta)\,f(I))\,]$.
The distribution $q(\Delta\theta)$ both draws and weights the training angles, so this
$q$-weighting is where pair-relative partial equivariance enters learning, concentrated
near $0^\circ$ for handheld imagery and near-uniform otherwise. For Exact and
PaSteer-soft the upright identity is exact by construction (\cref{eq:residual}), so the
$\Delta\theta=0$ term is guaranteed rather than learned.

\noindent\textbf{Joint objective.} In the full-scale joint regime the descriptor and
steerer train together under the standard DeDoDe matching loss \cite{edstedt2024dedode}
rather than the steering error, with continuous in-plane rotation augmentation that
exposes the steerer to all relative angles, not the multiples of $90^\circ$ used by the
original Steerers \cite{bokman2024steerers}; \cref{fig:angle} isolates this continuous
coverage. The joint comparison is reported in \cref{sec:pending}; the optimizer, step
count, learning rate, and data splits, shared across all steerers, are in
\cref{sec:experiments}.

\begin{table}[t]
  \vspace{-0.2cm}
  \centering\small
  \setlength{\tabcolsep}{5pt}
  \begin{tabular}{lccc}
    \toprule
    benchmark / metric  & Exact & PaSteer-free & $\Delta$ \\
    \midrule
    Roto-360 @5px        & 92.0 & \textbf{93.4} & $+1.4$ \\
    MD-1500 C4 AUC@5     & 32.5 & \textbf{36.4} & $+3.9$ \\
    MD-1500 SO2 AUC@5    & 20.0 & \textbf{22.9} & $+2.8$ \\
    AIMS AP              & 48.6 & \textbf{50.1} & $+1.5$ \\
    IMC-PT C4 AUC@20     & 67.2 & \textbf{75.2} & $+8.0$ \\
    IMC-PT SO2 AUC@20    & 41.1 & \textbf{50.6} & $+9.5$ \\
    \bottomrule
  \end{tabular}
  \vspace{-0.2cm}
  \caption{\textbf{Main result.} On frozen DeDoDe-B descriptors~\cite{edstedt2024dedode}, we measure absolute scores for the exact
  steerer and PaSteer-free with their difference $\Delta$, on Roto-360~\cite{lee2022rekd} correct matches @5px; MegaDepth-1500~\cite{li2018megadepth} pose
  AUC@5$^\circ$; AIMS~\cite{stoken2023aims} average precision on the full FMAP set (323 frames); IMC-PT~\cite{jin2021image} pose
  AUC@20$^\circ$. PaSteer-free improves over the exact steerer on all four benchmarks,  by the widest margin under large relative rotation.}
  \label{tab:margin}
\end{table}

\section{Experiments}
\label{sec:experiments}

We organize the experiments around five questions. \Cref{sec:matching} tests
whether relaxing the representation constraint improves descriptor matching across
benchmarks. \Cref{sec:theoryval} analyzes why the relaxation helps.
\Cref{sec:mechanisms} asks whether the gain comes merely from added capacity.
\Cref{sec:rotoreal} studies when equivariance matters under different
relative-rotation distributions. Finally, \Cref{sec:pending} clarifies the scope
of the relaxation under full joint descriptor--steerer training.

\begin{table}[t]
  \vspace{-0.2cm}
  \centering\small
  \setlength{\tabcolsep}{3pt}
  \begin{tabular}{lccccc}
    \toprule
    steerer  & Roto & upr. & C4 & SO2 & AIMS \\
    \midrule
    exact expm            & 92.0 & 49.0 & 32.5 & 20.0 & 48.6 \\
    PaSteer-free (ours)   & \textbf{93.4} & \textbf{49.3} & \textbf{36.4} & \textbf{22.9} & \textbf{50.1} \\
    \midrule
    $\Delta$ (free$-$exact) & $+1.4$ & $+0.3$ & $+3.9$ & $+2.8$ & $+1.5$ \\
    \bottomrule
  \end{tabular}
  \vspace{-0.2cm}
  \caption{\textbf{Full-scale MegaDepth training.} PaSteer-free versus the exact steerer,
  both trained on the full MegaDepth set on a frozen DeDoDe-B~\cite{edstedt2024dedode} with identical scenes,
  budget, generator core, and initialization. Roto denotes Roto-360 correct matches
  @5px~\cite{lee2022rekd}; upr., C4, SO2 denote MegaDepth-1500 pose AUC@5$^\circ$
  (upright/C4/SO2)~\cite{li2018megadepth}; AIMS denotes average precision on
  AIMS~\cite{stoken2023aims}. The $\Delta$ row is the relaxed-over-exact margin.}
  \label{tab:fullscale}
\end{table}

\paragraph{Model and steerer setup.}
Most matching results follow the Setting-A structure of
Steerers~\cite{bokman2024steerers}: the DeDoDe-B
descriptor~\cite{edstedt2024dedode} and the $\SOtwo$ detector are fixed, and only
the steering transform is trained. This setup isolates the representation
constraint, since the exact and PaSteer arms use the same descriptor, detector,
keypoints, matcher, initialization, and generator core, and differ only in the
steering form. We call this shared setup \emph{matched-budget}; parameter count is
not matched here and is controlled separately in \cref{sec:mechanisms}. Because
the descriptor itself is not adapted to rotation, this protocol also caps
absolute performance; the full-scale joint comparison is reported separately in
\cref{sec:pending}.

\paragraph{Training and matching details.}
Unless noted, all learned steerers are trained on a fixed
HPatches~\cite{balntas2017hpatches} training split for a few thousand Adam steps
and are evaluated with the same matching protocol: maximum-similarity matching
over $C_8$ probe angles, with dual-softmax~\cite{sun2021loftr} for the unsteered
baseline. For the closed-form optimal steerer of supp.\ Eq.~\ref{eq:optsteerer}, we estimate
$G$ and $C(\theta)$ on the same HPatches training split as the learned steerers
and report steering error on a disjoint HPatches evaluation split, using ridge
weight $\lambda=10^{-2}\cdot\mathrm{mean}(\mathrm{diag}(G))$. The HPatches train
and evaluation sequences, together with the Roto-360~\cite{lee2022rekd} test sequences, are listed
in supp.\ \cref{sec:supp-splits}.

\paragraph{Exact, PaSteer, and released Steerers.}
Our \emph{exact} baseline is the exact-representation counterpart to PaSteer, not
an ad-hoc baseline. It uses the same steering mechanism as
Steerers~\cite{bokman2024steerers}, a single $\SOtwo$ representation
$\rho(\theta)=\expmop(\theta D)$, under the same Setting-A structure. The
difference from the \emph{released} Steerers model is protocol rather than
mechanism: our exact and PaSteer variants use the matched-budget setup above,
whereas the released model is trained on MegaDepth~\cite{li2018megadepth} with rotations at multiples
of $90^\circ$ and uses a Procrustes matcher. The released Steerers rows in
\cref{tab:roto360} and supp.\ Tabs.~\ref{tab:md} and~\ref{tab:aims} are therefore absolute-performance references,
while the controlled comparison is exact versus PaSteer under matched training.

\paragraph{Datasets.}
We evaluate matching on Roto-360~\cite{lee2022rekd},
MegaDepth-1500~\cite{li2018megadepth}, AIMS~\cite{stoken2023aims}, and the
IMC-PT 2020 validation set~\cite{jin2021image}.

\begin{table}[t]
  \vspace{-0.2cm}
  \centering\small
  \setlength{\tabcolsep}{4pt}
  \scalebox{0.95}{
  \begin{tabular}{llccc}
    \toprule
     desc. & metric & Exact & PS-free & $\Delta$ \\
    \midrule
    \multirow{3}{*}{DeDoDe-B} & Roto-360 @5px     & 92.0 & 93.4 & $+1.4$ \\
                              & MD-1500 C4 AUC@5  & 32.5 & 36.4 & $+3.9$ \\
                              & MD-1500 SO2 AUC@5 & 20.0 & 22.9 & $+2.8$ \\
    \midrule
    \multirow{3}{*}{DeDoDe-G} & Roto-360 @5px     & 51.1 & \textbf{80.6} & $\mathbf{+29.5}$ \\
                              & MD-1500 C4 AUC@5  & 12.9 & \textbf{21.6} & $\mathbf{+8.7}$ \\
                              & MD-1500 SO2 AUC@5 & 11.8 & \textbf{20.1} & $\mathbf{+8.3}$ \\
    \bottomrule
  \end{tabular}
  }
  \vspace{-0.2cm}
  \caption{\textbf{The cost of the exact constraint grows with descriptor capacity.}
  Comparison with the exact steerer
  and PaSteer-free (PS-free) on DeDoDe-B~\cite{edstedt2024dedode} and the larger DeDoDe-G, on Roto-360~\cite{lee2022rekd} correct
  matches @5px and MegaDepth-1500~\cite{li2018megadepth} pose AUC@5$^\circ$ (C4/SO2). The margin widens markedly
  on DeDoDe-G across every rotation axis; scores are means over three training seeds
  (two for exact on DeDoDe-G).
  }
  \label{tab:scaleG}
\end{table}

\subsection{Does relaxation improve descriptor matching?}
\label{sec:matching}

Our central regime is frozen descriptors: equipping an already-trained descriptor with rotation robustness without retraining it, with the SO2
detector fixed.

\paragraph{The relaxed-over-exact margin.}
\Cref{tab:margin} is the main results: on a frozen descriptor, relaxing the representation
constraint improves matching on all four benchmarks, by the widest margin on the
large-rotation C4 and SO2 axes, while upright stays within a fraction of a point of
exact (slightly below for PaSteer-free, whose $\mathrm{err}(0)$ is near but nonzero, and
exactly zero for PaSteer-soft by construction). The reported $\Delta$ is a
controlled within-protocol difference that sidesteps the
AIMS~\cite{stoken2023aims} absolute-AP protocol gap noted below.

\paragraph{Full-scale Setting-A absolutes.}
\Cref{tab:fullscale} reports the full-scale MegaDepth~\cite{li2018megadepth} training results, where PaSteer-free wins on every axis. Additional results are in the supplement.

\paragraph{Scaling to a larger descriptor.}
\cref{tab:scaleG} shows the results on a larger-sized descriptor DeDoDe-G~\cite{edstedt2024dedode}.
The cost of enforcing an exact representation grows with descriptor capacity.
DeDoDe-G uses a substantially larger DINOv2 backbone than DeDoDe-B, and the
relaxed-over-exact margin widens on every rotation axis. This mirrors
\cref{sec:theoryval}: on the larger descriptor, the per-angle optimum departs
further from a group representation, a gap that only the free-form steerer can
effectively close.

\paragraph{Matched-budget three-way comparison.}
\Cref{tab:roto360} shows the Roto-360~\cite{lee2022rekd} matched-budget comparison among exact,
PaSteer-soft, and PaSteer-free, trained with the same data and budget on the same
HPatches~\cite{balntas2017hpatches} sequences. The ordering exact $<$ PaSteer-soft $<$ PaSteer-free holds, with
PaSteer-soft improving over the matched exact steerer while keeping
$\mathrm{err}(0)=0$ and PaSteer-free improving further. The released SO2-Spread-B
model reaches higher absolute accuracy under a different full-scale released
training protocol, so it should be read as an upper reference rather than a
controlled counterpart to PaSteer. The same matched-budget ordering reproduces on
MegaDepth-1500 and AIMS (PaSteer-soft $+7.3$ AP, $44.4$ vs.\ $37.1$; PaSteer-free
$51.2$), with full per-benchmark tables in supp.\ \cref{sec:supp-matched}.

\begin{table}[t]
  \vspace{-0.2cm}
  \centering\small
  \setlength{\tabcolsep}{4pt}
  \begin{tabular}{lccc}
    \toprule
    steering  & 3px & 5px & 10px \\
    \midrule
    \multicolumn{4}{c}{\textit{Matched-budget controlled comparison}} \\
    \midrule
    none (unsteered)             & 19.7 & 20.5 & 21.6 \\
    exact expm (our budget)      & 77.3 & 78.3 & 79.2 \\
    PaSteer-soft ($r{=}12$, ours)    & 89.1 & 90.1 & 91.0 \\
    PaSteer-free (ours)          & \textbf{93.9} & \textbf{95.1} & \textbf{96.0} \\
    \midrule
    \multicolumn{4}{c}{\textit{Released reference models, not controlled}} \\
    \midrule
    exact, release Setting-A & 91.4 & 93.0 & 94.1 \\
    Steerers SO2-Spread-B & 95.0 & 97.0 & 98.5 \\
    \bottomrule
  \end{tabular}
  \vspace{-0.2cm}
  \caption{\textbf{Matched-budget three-way comparison} on Roto-360~\cite{lee2022rekd}, correct matches
  \% @\{3,5,10\}px, frozen DeDoDe-B~\cite{edstedt2024dedode}. The top block is the matched-budget controlled
  comparison (ordering exact $<$ PaSteer-soft $<$ PaSteer-free); the bottom block lists
  released models as absolute references only, not controlled comparisons. Per-benchmark
  MegaDepth-1500 and AIMS tables are in supp.\ \cref{sec:supp-matched}.}
  \label{tab:roto360}
\end{table}

\begin{figure}[t]
  \vspace{-0.2cm}
  \centering
  \includegraphics[width=\linewidth]{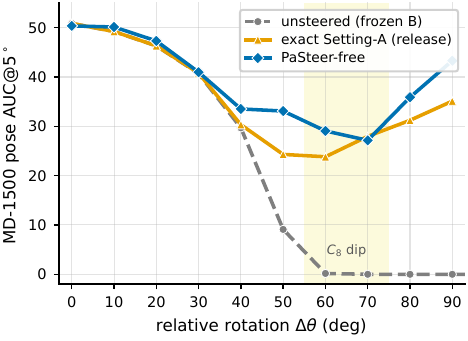}
  \vspace{-0.6cm}
  \caption{\textbf{Continuous-angle robustness.} Per-angle MD-1500~\cite{li2018megadepth} pose AUC@5$^\circ$
  as the relative in-plane rotation $\Delta\theta$ is swept continuously, on a frozen
  DeDoDe-B~\cite{edstedt2024dedode} descriptor with $C_8$ probes. \emph{Unsteered} (gray dashed) is the frozen
  descriptor without steering; \emph{exact Setting-A} (orange) is the released
  Steerers~\cite{bokman2024steerers} Setting-A model; \emph{PaSteer-free} (blue) is our
  relaxed steerer. PaSteer-free is at or above both across nearly the whole sweep, with
  the largest gains at the $C_8$ discretization-worst angles.
  }
  \label{fig:angle}
\end{figure}

\paragraph{Continuous-angle robustness.}
\Cref{fig:angle}  shows the results as pair relative rotation is swept continuously, isolating continuous
steering from augmenting only by multiples of $90^\circ$ on MD-1500 pose
AUC@5$^\circ$.  The unsteered descriptor is reliable only near upright rotations;
the exact steerer restores robustness but dips at the $C_8$ discretization-worst
angles. PaSteer-free leads across most angles, especially near these dips, and
also raises the accuracy level between probe angles, showing that the gain is
not just a $C_8$ gap-filling effect.

\subsection{Why does the partial equivariance help?}
\label{sec:theoryval}

\Cref{tab:theory} shows steering-error results on HPatches~\cite{balntas2017hpatches} evaluation set using
a frozen DeDoDe-B~\cite{edstedt2024dedode} descriptor. We compare each learned steerer with the
closed-form per-angle optimum $\rho^\star$ of supp.\ Eq.~\ref{eq:optsteerer}, which serves
as an oracle lower-error reference rather than a deployable model.

The exact steerer is consistently farther from this optimum, especially at large
rotations. PaSteer-free stays close to $\rho^\star$ at all reported angles, and
PaSteer-soft falls in between (\cref{fig:floor}). The bottom rows show that
$\rho^\star$ has a large composition violation, meaning that the best
angle-wise maps do not compose as a single $\SOtwo$ representation. This
explains why the exact representation constraint is limiting for a frozen
descriptor, and why PaSteer-free can better match its actual angular response.

\begin{table}[t]
  \vspace{-0.2cm}
  \centering\small
  \setlength{\tabcolsep}{3pt}
  \begin{tabular}{lcccc}
    \toprule
    $\theta$ & $\rho^\star$ & exact & PaSteer-soft & PaSteer-free \\
    \midrule
    $30^\circ$  & 0.229 & 0.305 & 0.311 & \textbf{0.239} \\
    $90^\circ$  & 0.264 & 0.360 & 0.348 & \textbf{0.280} \\
    $180^\circ$ & 0.236 & 0.452 & 0.366 & \textbf{0.248} \\
    \midrule
    comp.\ violation    & \multicolumn{4}{c}{$0.60$--$0.70$} \\
    orthog.\ deviation  & \multicolumn{4}{c}{$0.85$--$1.04$} \\
    \bottomrule
  \end{tabular}
  \vspace{-0.2cm}
  \caption{
  \textbf{Steering error against the per-angle optimum.}
Steering error $1-\cos$ is reported on HPatches~\cite{balntas2017hpatches} evaluation set using a frozen
DeDoDe-B~\cite{edstedt2024dedode} descriptor. The top rows compare the closed-form
per-angle optimum $\rho^\star$, the exact steerer, PaSteer-soft, and
PaSteer-free at selected rotation angles. The bottom rows report the
composition violation and orthogonality deviation of $\rho^\star$: the former
measures how far $\rho^\star(\theta+\phi)$ is from
$\rho^\star(\theta)\rho^\star(\phi)$, and the latter measures how far
$\rho^{\star\top}\rho^\star$ is from the identity.
  }
  \label{tab:theory}
\end{table}

\subsection{Capacity and constraint controls}
\label{sec:mechanisms}

\paragraph{Mechanism comparison.}
\cref{tab:mechanism} shows that PaSteer-free obtains the lowest steering error at every angle but at a large parameter
cost, raising whether the extra capacity, not the relaxation,
buys the gain. To separate the two we use two controls: \emph{PaSteer-soft}, a
capacity-light residual that keeps $\mathrm{err}(0)=0$ and whose out-of-band gain over
exact \emph{saturates} with rank, so it is a functional-form effect rather than added
capacity; and an \emph{orthogonality-preserving residual} of matched size that isolates
the orthogonality constraint. \Cref{tab:mechanism} reports steering error in and out of
the natural partial-equivariance band of \cref{fig:angle}.

\begin{table}[t]
  \vspace{-0.2cm}
  \centering\small
  \setlength{\tabcolsep}{4pt}
  \begin{tabular}{lcccc}
    \toprule
    config & +params\% & in-band & out-band & err(0) \\
    \midrule
    exact expm              & $0$     & 0.152 & 0.410 & 0.000 \\
    PaSteer-soft ($r{=}12$)  & $+75$   & 0.156 & \textbf{0.374} & 0.000 \\
    PaSteer-soft ($r{=}24$)  & $+150$  & 0.155 & 0.379 & 0.000 \\
    PaSteer-free            & $+1200$ & \textbf{0.123} & \textbf{0.297} & 0.006 \\
    \bottomrule
  \end{tabular}
  \vspace{-0.2cm}
  \caption{\textbf{The gain is functional form, not parameter count.} Steerer mechanism
  comparison on a frozen DeDoDe-B~\cite{edstedt2024dedode} descriptor (HPatches~\cite{balntas2017hpatches} evaluation set): steering error
  $1-\cos$ in and out of the partial-equivariance band, with err(0) the error at $0^\circ$.
  The out-of-band gain of PaSteer-soft over exact saturates with rank despite added parameters,
  so the remaining gap to PaSteer-free reflects leaving the representation and orthogonality
  constraints, not capacity.}
  \label{tab:mechanism}
\end{table}

\paragraph{Deployability.}
PaSteer-free is fully deployable: as in
\cref{sec:steering}, all three steerers are precomputed linear maps
over the shared $C_8$ probe grid scored by the same matcher, so PaSteer-free runs at
exactly the inference cost of the exact steerer, its larger parameterization a one-time
training and storage cost, not a per-match one. The $+1200\%$ is relative to the small exact
generator; PaSteer-free has $0.85$M parameters, $6.3\%$ of the DeDoDe-B descriptor.

\paragraph{Capacity fairness.}
\Cref{tab:ksweep} shows a sweep of the harmonic order $K$ of the free-form Fourier map
that separates relaxation from raw capacity; the parameter cost $(2K{+}1)\cdot 256^2$ grows
directly with $K$.
More parameters alone do not help: $K{=}1$ already has more parameters than the exact steerer
($1.5\%$ vs.\ $0.48\%$ of the descriptor) yet is markedly worse, and the relaxed-over-exact
margin turns positive from $K{=}3$ on, so $K{\ge}3$ is a safe default (we use $K{=}6$). The gain thus comes
from leaving the representation and orthogonality constraints, not from parameter count.

\begin{table}[t]
  \vspace{-0.2cm}
  \centering
  \footnotesize
  \setlength{\tabcolsep}{4pt}
  \begin{tabular}{lcccccc}
    \toprule
    steerer & params (\%) & Roto@5 & C4@5 & SO2@5 & $\Delta$C4 & $\Delta$SO2 \\
    \midrule
    exact   & 0.48 & 92.1 & 32.3 & 20.3 & -- & -- \\
    $K{=}1$ & 1.5  & 89.0 & 20.8 & 12.4 & $-11.6$ & $-7.9$ \\
    $K{=}2$ & 2.4  & 92.2 & 29.9 & 19.2 & $-2.4$ & $-1.0$ \\
    $K{=}3$ & 3.4  & 93.1 & 34.6 & 22.6 & $+2.3$ & $+2.3$ \\
    $K{=}4$ & 4.4  & 93.4 & 35.1 & 23.5 & $+2.7$ & $+3.2$ \\
    $K{=}6$ & 6.3  & \textbf{93.6} & \textbf{37.9} & \textbf{23.9} & $\mathbf{+5.6}$ & $\mathbf{+3.7}$ \\
    \bottomrule
  \end{tabular}
  \vspace{-0.2cm}
  \caption{\textbf{Free-Fourier capacity sweep over harmonic order $K$.} Frozen DeDoDe-B~\cite{edstedt2024dedode}
  Setting-A: Roto@5 is Roto-360~\cite{lee2022rekd} correct matches \% @5px; C4@5 and SO2@5 are
  MegaDepth-1500~\cite{li2018megadepth} pose AUC@5$^\circ$; $\Delta$C4 and $\Delta$SO2 are PaSteer-free$(K)$
  minus exact; params (\%) is steerer parameters as a fraction of the descriptor. More
  parameters alone do not beat exact: a sufficient harmonic order in the free-form
  parameterization is needed to cross it.}
  \label{tab:ksweep}
\end{table}

\subsection{When does equivariance matter?}
\label{sec:rotoreal}

\Cref{tab:rotoreal} shows when equivariance matters by changing the
pair-relative rotation distribution $q(\Delta\theta)$ while keeping the
matching setup fixed. We evaluate RotoReal by resampling MD-1500~\cite{li2018megadepth} pairs under
two distributions: T1 draws $|\Delta\theta|$ from a measured handheld
distribution, whereas T3 samples rotations uniformly over $\SOtwo$.

The result supports the pair-relative view of \Cref{def:pairrel}. On T1, the
common small-rotation regime, steering is nearly irrelevant because the upright
descriptor already covers the rotations realized in the data. On T3, where large
rotations are common, the methods separate clearly and the matched-training
ordering exact $<$ PaSteer-soft $<$ PaSteer-free reappears.

This is the empirical core of partial equivariance: the benefit of full
$\SOtwo$ equivariance is small when $q(\Delta\theta)$ is concentrated near
upright, but becomes important when $q$ places mass on large rotations. Thus, the
useful degree of equivariance is not fixed globally; it is determined by the
pair-relative rotation distribution of the domain.

\begin{table}[t]
  \vspace{-0.2cm}
  \centering\small
  \begin{tabular}{lcc}
    \toprule
    steering  & T1 handheld & T3 uniform \\
    \midrule
    dual softmax (none)         & 50.8 & 10.7 \\
    exact expm (ours)           & 50.7 & 19.7 \\
    PaSteer-soft ($r{=}12$, ours)   & 50.7 & 24.7 \\
    PaSteer-free (ours)         & 49.9 & \textbf{35.0} \\
    \midrule
    \emph{Steerers Setting-A (release, ref.)} & \emph{50.7} & \emph{30.6} \\
    \bottomrule
  \end{tabular}
  \vspace{-0.2cm}
  \caption{\textbf{The useful degree of equivariance depends on $q(\Delta\theta)$.}
  RotoReal MD-1500~\cite{li2018megadepth} pose AUC@5$^\circ$ on a frozen DeDoDe-B~\cite{edstedt2024dedode}, resampling pairs under a
  measured handheld distribution (T1) versus uniform over $\SOtwo$ (T3); only the steering
  changes. Steering is near-irrelevant on T1 and separates the methods only on T3. The
  release row is MegaDepth-trained and shown for reference.}
  \label{tab:rotoreal}
\end{table}

\subsection{What is the scope?}
\label{sec:pending}

\Cref{tab:pending} shows the full-scale joint descriptor--steerer comparison on
MegaDepth~\cite{li2018megadepth}, where only the steerer mechanism differs. Exact and PaSteer-free reach
parity, suggesting that joint training lets the descriptor absorb the
representation constraint.
Thus, PaSteer is primarily a frozen-descriptor retrofit method: it helps most
when the descriptor cannot be retrained, while becoming largely neutral under
full joint training. The larger DeDoDe-G~\cite{edstedt2024dedode} joint result shows the same scope
boundary, with a small edge for exact steering (supp.\ \cref{sec:supp-gjoint}).

\begin{table}[t]
  \vspace{-0.2cm}
  \centering\small
  \begin{tabular}{lccc}
    \toprule
    model  & MD upright & MD C4 & MD SO2 \\
    \midrule
    exact steerer             & \textbf{50.3} & 47.0 & 43.3 \\
    PaSteer-free (ours)       & 49.7 & \textbf{47.1} & \textbf{43.6} \\
    \bottomrule
  \end{tabular}
  \vspace{-0.2cm}
  \caption{\textbf{Joint training reaches parity, isolating the frozen-descriptor
  advantage.} Full-scale joint descriptor--steerer training on DeDoDe-B~\cite{edstedt2024dedode}, pose
  AUC@5$^\circ$ on MegaDepth-1500~\cite{li2018megadepth}, with identical scenes, schedule, generator core, and
  initialization so only the steerer mechanism differs. Exact and PaSteer-free reach
  parity once the descriptor co-adapts, so the advantage of the relaxation is concentrated in
  the frozen-descriptor regime (supp.\ Tab.~\ref{tab:md}).}
  \label{tab:pending}
\end{table}

\section{Conclusion}
\label{sec:conclusion}

We formulated rotation-robust keypoint descriptor matching as pair-relative
partial equivariance, where equivariance is measured against the empirical
relative-rotation distribution $q(\Delta\theta)$ rather than the uniform group.
For a fixed, non-equivariant descriptor, the per-angle optimal linear steerer is
generally not a group representation, which explains why an exact $\SOtwo$
representation can be restrictive in the frozen-descriptor regime. PaSteer
relaxes this constraint with two variants: PaSteer-soft, which preserves the
upright identity exactly, and PaSteer-free, which directly parameterizes the
angle-dependent map.

Our experiments support this view. The steering-error analysis shows that the
best steering map changes with rotation angle and is not well captured by one
exact representation. Matching experiments show that PaSteer improves robustness
to large rotations while preserving upright performance, and distribution
controlled experiments confirm that steering matters most when large relative
rotations are common. Overall, PaSteer provides a lightweight way to add rotation
robustness to already-trained descriptors without changing the descriptor
backbone.

A natural next step is to extend this pair-relative, data-driven view from in-plane rotation to richer geometric transformations such as scale, affine, and viewpoint changes, potentially reducing the amount of augmentation or training data needed to obtain robust descriptors.

\paragraph{Limitations.}
Our evaluation is limited to DeDoDe descriptors, and we compare against exact steering
rather than rotation-equivariant methods that retrain the backbone. The steerer also
applies the same $C_8$ grid to every pair instead of adapting its degree of equivariance.

\paragraph{Acknowledgements.}
This work was supported by the National Research Foundation of Korea (NRF) grant funded by the Korea government (MSIT) (RS-2026-25598824).

{
    \small
    \bibliographystyle{ieeenat_fullname}
    \bibliography{main}
}

\clearpage
\appendix
\begin{center}
  {\Large\bfseries Pair-Relative Partial Equivariance for Rotation-Robust\\
  Keypoint Descriptor Matching\par}
  \vspace{0.7em}
  {\large --- Supplementary Material ---\par}
\end{center}
\label{sec:supp}
\vspace{0.8em}

This supplement collects analyses that support the main paper but do not fit its page
budget. \Cref{sec:supp-theory} develops the full theory (the closed-form optimal steerer
and the non-homomorphism theorem). \Cref{sec:supp-matched} reports the matched-budget
per-benchmark absolute scores for the three-way comparison. \Cref{sec:supp-aims} details
the AIMS~\cite{stoken2023aims} evaluation protocol. \Cref{sec:supp-gjoint} delimits the scope under DeDoDe-G~\cite{edstedt2024dedode}
joint training. \Cref{sec:supp-splits} lists the HPatches~\cite{balntas2017hpatches} training, evaluation, and test
splits. \Cref{sec:supp-qual} shows dataset-by-dataset qualitative comparisons. Throughout,
\textbf{PaSteer-free} is our free-form Fourier steerer and \emph{exact} is the Steerers~\cite{bokman2024steerers}
representation retrained at our budget on a frozen DeDoDe-B descriptor.

\section{Why relaxation helps: theory}
\label{sec:supp-theory}

This section gives the theory behind PaSteer in full, expanding the brief summary
in \cref{sec:pairrel}. The argument is short and has one moving part: for a
\emph{fixed}, non-equivariant descriptor, the best possible linear steerer at each
angle is itself generally not a group representation, so an exact representation
steerer is provably suboptimal and a relaxed steerer can do strictly better.

\paragraph{The best steerer at a single angle.}
Fix the descriptor $f$ and one rotation angle $\theta$. The steerer that best
transports descriptors from the upright frame into the $\theta$-rotated frame is
the linear map minimizing the average squared steering error,
\begin{equation}
\rho^\star(\theta)=\argmin_{M\in\R^{D\times D}}\;\E_I\big\|f(R_\theta I)-M f(I)\big\|^2 .
\label{eq:perangle}
\end{equation}
This is an ordinary ridge least-squares problem, so it has the closed-form
solution
\begin{equation}
\begin{aligned}
  \rho^\star(\theta)
  &= C(\theta)\,(G+\lambda I)^{-1}, \\
  C(\theta)
  &= \E\!\left[f(R_\theta I)f(I)^\top\right], \\
  G
  &= \E\!\left[f(I)f(I)^\top\right].
\end{aligned}
\label{eq:optsteerer}
\end{equation}
Here $G$ is the covariance of the descriptor and $C(\theta)$ is the
cross-covariance between upright and rotated descriptors. We call
$\rho^\star$ the \emph{per-angle optimal steerer}: it is the best any linear
steerer can do at angle $\theta$, but it estimates a separate full
$D\times D$ matrix for each angle and therefore serves as an oracle reference
rather than a practical compact model.

\paragraph{The ceiling is not a representation.}
The exact steerer implicitly assumes that the angle-wise transport maps chain
together into a representation, $\rho(\theta+\phi)=\rho(\theta)\rho(\phi)$.
This holds only under a restrictive second-order condition. Write
$L(\rho)=\E_{\theta\sim q}\,\E_I\,\|f(R_\theta I)-\rho(\theta)f(I)\|^2$ for the
aggregate steering loss.

\begin{theorem}[The per-angle optimum is generally not a representation]
\label{thm:A}
Assume $\lambda\to0^+$ and $G$ is invertible, so that
$\rho^\star(\theta)=C(\theta)G^{-1}$. Then $\rho^\star$ satisfies
$\rho^\star(\theta+\phi)=\rho^\star(\theta)\rho^\star(\phi)$ for all
$\theta,\phi$ if and only if
\begin{equation}
  C(\theta+\phi)=C(\theta)G^{-1}C(\phi).
  \label{eq:comp-condition}
\end{equation}
\end{theorem}

Thus, the per-angle optimum is a representation only when the angle-wise
descriptor correlations compose consistently. Frozen upright-trained descriptors
generally violate this condition, so their optimal angle-wise steerers need not
form a single exact $\SOtwo$ action. Because $\rho^\star$ is the angle-wise
least-squares optimum, it lower-bounds the loss of any exact representation, with
a strict gap whenever the least-squares optimum is unique and the violation occurs
on angles of positive $q$-measure. This is the load-bearing claim of the paper.

\paragraph{Two caveats.}
First, failure of the \emph{composition law} is the property we rely on. A
separate, weaker fact is that $\rho^\star$ need not be \emph{orthogonal}; but an
$\SOtwo$ representation need not be orthogonal either, unless one additionally
asks for a unitary representation. We therefore report both metrics and lean on
the composition violation. Second, at finite ridge $\lambda>0$ the solution has
$\rho^\star(0)=G(G+\lambda I)^{-1}\neq I$. This is a numerical regularization
effect rather than a property of $f$, and the zero-angle error vanishes in the
unregularized limit $\lambda\to0^+$. We therefore state the theorem in the clean
population limit and use $\lambda$ only to stabilize the estimator.

\paragraph{From ceiling to method.}
$\rho^\star$ is an oracle per-angle matrix; the practical question is how closely
a compact steerer can approach it. PaSteer-free approximates $\rho^\star$ with a
handful of harmonics, while PaSteer-soft (\cref{eq:relaxed}) trades a little of
this gap for an exact upright identity $\rhotil(0)=I$ and graceful extrapolation
beyond the training angles. \Cref{sec:theoryval} verifies all of this on real
DeDoDe-B~\cite{edstedt2024dedode} and DeDoDe-G descriptors: the per-angle
optimum violates the composition law, a free-form steerer nearly attains the
optimal floor, and a trained exact steerer falls short, with the gap widening on
the larger descriptor, as the theorem predicts.

\section{Matched-budget per-benchmark absolutes}
\label{sec:supp-matched}

\paragraph{Controlled-comparison convention.}
Unless stated otherwise, a controlled comparison means that the descriptor,
detector, keypoints, matcher, data split, training budget, and evaluation protocol
are fixed, and only the steerer changes. Released or published models are included
only as absolute-performance references and are not used for mechanism-level claims.

\Cref{tab:md,tab:aims} show the MegaDepth-1500~\cite{li2018megadepth} and AIMS~\cite{stoken2023aims} per-benchmark absolute scores
behind the matched-budget three-way comparison (exact, PaSteer-soft, PaSteer-free) of
\cref{sec:matching}, each trained on the same HPatches~\cite{balntas2017hpatches} sequences over a frozen DeDoDe-B~\cite{edstedt2024dedode}
descriptor; the Roto-360~\cite{lee2022rekd} table is in the main paper (\cref{tab:roto360}). These complement
the main comparison by confirming the same ordering, exact $<$ PaSteer-soft $<$
PaSteer-free, on per-benchmark pose AUC and average precision. Following the convention
above, the top block of each table is the controlled comparison and the bottom block lists
released models as absolute references only, so the mechanism-level claim should be read
from the matched-budget block. The released exact steerer is MegaDepth-trained (hence
in-domain on MegaDepth-1500) and the released joint Steerers~\cite{bokman2024steerers} additionally differ in
steerer family and, on AIMS, use a non-public Procrustes protocol (\cref{sec:supp-aims}).

\begin{table}[t]
  \centering\small
  \setlength{\tabcolsep}{4pt}
  \begin{tabular}{lccc}
    \toprule
    steering  & upright & C4 & SO2 \\
    \midrule
    \multicolumn{4}{c}{\textit{Matched-budget controlled comparison}} \\
    \midrule
    dual softmax (none)            & 49.8 & 12.4 & 11.4 \\
    exact expm (ours, HPatches)    & 49.8 & 28.5 & 19.4 \\
    PaSteer-soft ($r{=}12$, ours)      & 49.8 & 37.5 & 24.4 \\
    PaSteer-free (ours)            & 49.0 & \textbf{43.1} & \textbf{33.0} \\
    \midrule
    \multicolumn{4}{c}{\textit{Released reference models, not controlled}} \\
    \midrule
    exact Setting-A (release)      & 49.8 & 39.5 & 31.3 \\
    Steerers SO2-Spread-B & 49.7 & 49.0 & 47.5 \\
    \bottomrule
  \end{tabular}
  \caption{\textbf{MegaDepth-1500~\cite{li2018megadepth} matched-budget absolutes.} Pose AUC@5$^\circ$ on a
  frozen DeDoDe-B~\cite{edstedt2024dedode}. The top block is the matched-budget controlled comparison
  (exact $<$ PaSteer-soft $<$ PaSteer-free); the bottom block lists released models as
  absolute references only.}
  \label{tab:md}
\end{table}

\begin{table}[t]
  \centering\small
  \begin{tabular}{lc}
    \toprule
    steering  & AP \\
    \midrule
    \multicolumn{2}{c}{\textit{Matched-budget controlled comparison}} \\
    \midrule
    unsteered (dual softmax)             & 25.5 \\
    exact expm (ours, HPatches)          & 37.1 \\
    PaSteer-soft ($r{=}12$, ours)            & 44.4 \\
    PaSteer-free (ours)                  & \textbf{51.2} \\
    \midrule
    \multicolumn{2}{c}{\textit{Released reference models, not controlled}} \\
    \midrule
    exact Setting-A (release, MegaDepth) & 46.3 \\
    Steerers SO2-Freq1-B & 57.7 \\
    \bottomrule
  \end{tabular}
  \caption{\textbf{AIMS~\cite{stoken2023aims} matched-budget absolutes.} Average precision on the FMAP
  evaluation subset, frozen DeDoDe-B~\cite{edstedt2024dedode}, with identical candidate set and positives across
  rows. The top block is the matched-budget controlled comparison (same ordering as
  Roto-360~\cite{lee2022rekd}); the bottom block lists released models as absolute references only.}
  \label{tab:aims}
\end{table}

\subsection{Additional full-scale absolute results and protocol notes}
\label{sec:supp-fullscale-extra}

\Cref{tab:imc,tab:supp-aims} show full-scale Setting-A controlled comparisons of the
exact and relaxed steerers on IMC-PT~\cite{jin2021image} and the AIMS~\cite{stoken2023aims} full-FMAP set, on a frozen DeDoDe-B~\cite{edstedt2024dedode}. The AIMS numbers in \cref{tab:aims} and \cref{tab:supp-aims} correspond to different training scales: HPatches matched-budget versus full-scale Setting-A.
These complement the headline margins (\cref{tab:margin}) by giving the absolute scores
behind them at full scale: PaSteer-free improves over exact mainly on the large-rotation
axes and splits while staying within a fraction of a point of exact near upright. Protocol
caveats specific to AIMS are deferred to \cref{sec:supp-aims}.

\begin{table}[t]
  \centering\small
  \setlength{\tabcolsep}{3pt}
  \begin{tabular}{lccc}
    \toprule
    IMC-PT~\cite{jin2021image} & upright & C4 & SO2 \\
    \midrule
    exact            & 66.4/80.7/89.5 & 41.5/55.6/67.2 & 22.2/31.3/41.1 \\
    PaSteer-free    & 66.0/80.6/89.6 & \textbf{47.3/62.7/75.2} & \textbf{28.7/40.1/50.6} \\
    \bottomrule
  \end{tabular}
  \caption{\textbf{IMC-PT full-scale Setting-A controlled comparison.} IMC-PT 2020~\cite{jin2021image}
  validation (1500 pairs/axis), frozen DeDoDe-B~\cite{edstedt2024dedode}, pose AUC@\{5,10,20\}$^\circ$. PaSteer-free
  improves over exact on the large-rotation C4/SO2 axes (up to $+8.0$/$+9.5$ AUC) while
  staying within a fraction of a point of exact on upright.}
  \label{tab:imc}
\end{table}

\Cref{tab:supp-aims} reports the controlled AIMS full-FMAP comparison: the exact and
relaxed steerers share the frozen DeDoDe-B descriptor, candidate set, positives, $C_8$
maximum-similarity matcher, and training budget, with only the steerer mechanism differing.
PaSteer-free improves over exact on both the overall and the large-rotation (A.O.) split.
The absolute level
reflects the AIMS protocol of \cref{sec:supp-aims}.

\begin{table}[t]
  \centering\small
  \setlength{\tabcolsep}{6pt}
  \begin{tabular}{lcc}
    \toprule
    steerer & AP All & AP A.O. \\
    \midrule
    exact expm            & 48.63 & 43.40 \\
    PaSteer-free (ours)        & \textbf{50.11} & \textbf{45.24} \\
    \midrule
    margin                & $+1.48$ & $+1.85$ \\
    \bottomrule
  \end{tabular}
  \caption{\textbf{AIMS~\cite{stoken2023aims} full-FMAP controlled comparison.} Exact versus PaSteer-free on the
  full FMAP evaluation set ($323$ frames, $21\,777$ pairs), frozen DeDoDe-B~\cite{edstedt2024dedode}, full-scale
  Setting-A, $C_8$ maximum-similarity matching with identical candidate set and positives.
  AP All is over all pairs and AP A.O.\ is the large-rotation (any-orientation) split;
  PaSteer-free improves on both. Protocol caveats are in \cref{sec:supp-aims}.}
  \label{tab:supp-aims}
\end{table}

\section{AIMS evaluation protocol}
\label{sec:supp-aims}

This section explains why our AIMS average precision differs from the published Steerers
number and why the within-protocol exact-vs-PaSteer comparison is the relevant evidence.
The AIMS \cite{stoken2023aims} average precision we report on a frozen descriptor
(Setting-A) is in the $46$--$50$\,AP range, whereas Steerers \cite{bokman2024steerers}
report $\approx 60$\,AP. The two numbers are produced by different evaluation protocols
and are not directly comparable: they differ in the matcher, the candidate set, and the
positive-label protocol. The reported $\approx 60$\,AP uses a Procrustes matcher with the
released, MegaDepth-trained~\cite{li2018megadepth} SO2-Freq1-B model and a candidate retrieval set and
positive-label margin that are not public. Our controlled study instead fixes the public
FMAP candidate and positive set and uses maximum-similarity matching over $C_8$ probe
angles for every method, the protocol under which all steerers in this paper are compared
on equal footing. Under this protocol the released exact Setting-A steerer scores
$\approx 46$\,AP ($44.9$ on the $162$-frame subset, $46.3$ on the full set), which is the
within-protocol anchor; under the Procrustes and full-FMAP protocol the same pipeline
reproduces $57.7$\,AP, matching the published setting. The relevant quantity is therefore
the within-protocol exact-vs-PaSteer-free comparison (\cref{tab:supp-aims}).

\section{Joint training on DeDoDe-G}
\label{sec:supp-gjoint}

\Cref{tab:supp-gjoint} shows full-scale joint descriptor--steerer training on the larger
DeDoDe-G~\cite{edstedt2024dedode} descriptor (DINOv2 ViT-L backbone), with identical scenes, schedule, generator
core, and initialization so that only the steerer mechanism differs; the exact generator
is held skew-symmetric to remain a valid representation. It complements the DeDoDe-B joint
result of the main paper (\cref{sec:pending}), where the two arms reach parity once the
descriptor co-adapts.

\begin{table}[t]
  \centering\small
  \setlength{\tabcolsep}{3pt}
  \scalebox{0.9}{
  \begin{tabular}{lcccc}
    \toprule
    model  & Roto & MD upr. & MD C4 & MD SO2 \\
    \midrule
    exact steerer (joint)            & \textbf{91.9} & \textbf{49.7} & \textbf{46.1} & \textbf{42.3} \\
    PaSteer-free (ours, joint)            & 90.7 & 45.8 & 41.5 & 38.2 \\
    PaSteer-free (pinned, ours, joint)    & --   & 47.7 & 42.2 & 39.2 \\
    \bottomrule
  \end{tabular}
  }
  \caption{\textbf{Joint training on DeDoDe-G.}
Full-scale joint descriptor--steerer training on DeDoDe-G~\cite{edstedt2024dedode}.
We report Roto-360~\cite{lee2022rekd} correct matches @5px and
MegaDepth-1500~\cite{li2018megadepth} pose AUC@5$^\circ$ for upright, C4, and
SO2 rotations. All rows use the same scenes, schedule, generator core, and
initialization; only the steerer mechanism differs. Unlike the frozen-descriptor
setting, the exact steerer keeps a small edge after the descriptor co-adapts,
while pinning $\tilde{\rho}(0)=I$ recovers part of the PaSteer-free upright gap.}
  \label{tab:supp-gjoint}
\end{table}

Under joint training on the larger G descriptor the exact steerer marginally
\emph{outperforms} PaSteer-free on the rotation axes (C4, SO2) and on upright; pinning the
upright identity $\rhotil(0)=I$ of the relaxed steerer (the PaSteer-free (pinned) row, two
seeds) recovers about half of the gap but does not reverse it. This is the regime opposite
to the frozen DeDoDe-G Setting-A result, where the relaxed steerer dominates the exact one
by a wide margin on every rotation axis (\cref{tab:scaleG}). This result delimits the
scope of PaSteer: relaxation helps in the frozen-descriptor retrofit regime, whereas a
jointly trained descriptor can adapt to, or even benefit from, the exact representation
constraint.

\section{HPatches training, evaluation, and test splits}
\label{sec:supp-splits}

Roto-360~\cite{lee2022rekd} applies in-plane rotations to the $116$
HPatches~\cite{balntas2017hpatches} sequences (prefix \texttt{i\_} for illumination
sequences, \texttt{v\_} for viewpoint sequences). For the matched-budget steerers we do
not introduce a new evaluation boundary: the $10$ sequences of the REKD/RELF Roto-360
\texttt{debug} split~\cite{lee2022rekd,lee2023relf} are used verbatim as the Roto-360
matching test set and never enter training or evaluation. The remaining non-\texttt{debug}
sequences are placed in a fixed order and shuffled once with a fixed random seed; the first
$60$ form the steerer training set and the next $12$ the steering-error evaluation set,
leaving the remaining $34$ unused. We fix the seed so that this partition is deterministic
and identical for every steerer variant: the exact and relaxed arms are then trained and
evaluated on exactly the same sequences, so any difference between them reflects the
steering form alone rather than a different split. This $60/12$ partition is newly defined
here.

These HPatches splits govern only the matched-budget, HPatches-trained steerers and the
steering-error and theory results: \cref{tab:mechanism,tab:theory} and the matched blocks
of \cref{tab:roto360,tab:md,tab:aims} (the ``ours, HPatches'' rows). The full-scale
Setting-A results (\cref{tab:fullscale,tab:imc,tab:scaleG}) and the joint training of
\cref{tab:pending,tab:supp-gjoint} are trained on MegaDepth and do not use this partition.

\smallskip
\noindent\textbf{Train ($60$).}
{\small\ttfamily i\_books, i\_boutique, i\_bridger, i\_brooklyn, i\_castle, i\_chestnuts,
i\_crownday, i\_crownnight, i\_dc, i\_dome, i\_fenis, i\_fruits, i\_gonnenberg,
i\_greenhouse, i\_greentea, i\_indiana, i\_kions, i\_kurhaus, i\_leuven, i\_lionday,
i\_lionnight, i\_melon, i\_miniature, i\_nijmegen, i\_partyfood, i\_pencils, i\_pinard,
i\_pool, i\_salon, i\_ski, i\_smurf, i\_steps, i\_yellowtent, v\_abstract, v\_adam,
v\_artisans, v\_boat, v\_bricks, v\_calder, v\_cartooncity, v\_churchill, v\_circus,
v\_coffeehouse, v\_dirtywall, v\_dogman, v\_graffiti, v\_home, v\_laptop, v\_machines,
v\_maskedman, v\_pomegranate, v\_posters, v\_soldiers, v\_strand, v\_sunseason,
v\_underground, v\_vitro, v\_wapping, v\_war, v\_wounded.}

\smallskip
\noindent\textbf{Evaluation set for steering error ($12$).}
{\small\ttfamily i\_autannes, i\_ktirio, i\_londonbridge, i\_tools, i\_veggies, i\_zion,
v\_azzola, v\_beyus, v\_bird, v\_busstop, v\_man, v\_samples.}

\smallskip
\noindent\textbf{Roto-360 test (the REKD/RELF \texttt{debug} split, $10$).}
{\small\ttfamily i\_ajuntament, i\_bologna, i\_resort, i\_table, i\_troulos, v\_birdwoman,
v\_courses, v\_gardens, v\_tabletop, v\_wormhole.}

\section{Qualitative comparison}
\label{sec:supp-qual}

\Cref{fig:qual-md,fig:qual-imc,fig:qual-roto,fig:qual-aims} show qualitative comparisons
on MegaDepth-1500~\cite{li2018megadepth}, IMC-PT~\cite{jin2021image}, Roto-360~\cite{lee2022rekd}, and AIMS~\cite{stoken2023aims}, illustrating dataset by dataset what the
relaxed steerer recovers that an exact representation cannot. Each figure compares three
steerers on one image pair, all on the
same frozen DeDoDe-B~\cite{edstedt2024dedode} descriptor, the same SO2 detector, and the same keypoints, so any
difference is attributable to the steerer alone: \textbf{(a)~no steer} (the raw upright
descriptor), \textbf{(b)~the exact Steerers~\cite{bokman2024steerers} representation} $\rho(\theta)=\expmop(\theta
D)$, and \textbf{(c)~PaSteer-free}. The color convention is fixed throughout: green lines
are ground-truth-verified correct matches, blue lines (panel~(c)) are the correct matches
PaSteer-free recovers but the exact steerer misses, and faint orange lines are rejected
matches (subsampled); the number of inlier lines drawn is capped for legibility while the
panel titles report the true counts. We rotate the synthetic-rotation pairs
(MegaDepth-1500, IMC-PT) to $112^\circ$ and the Roto-360 pair to $70^\circ$, angles off
the $C_8$ probe grid $\{0^\circ,45^\circ,\dots\}$ (the ``$C_8$-dip'' angles of
\cref{fig:angle}) where the exact representation is weakest; AIMS pairs carry their own
intrinsic rotation and are not rotated synthetically.

Across all four datasets the pattern is the same: without steering the upright descriptor
collapses under large rotation, the exact steerer recovers most of the structure, and
PaSteer-free recovers a further set of correct matches (blue) that the representation
constraint leaves unmatched. The effect is largest on AIMS, where the pair-relative
rotation distribution $q(\Delta\theta)$ is closest to uniform (\cref{sec:rotoreal}) and the
cost of forcing a single exact representation over all of $\SOtwo$ is highest. The
qualitative pattern is consistent with the steering-error and non-homomorphism analyses in
\cref{sec:supp-theory} and \cref{sec:theoryval}.

\twofigpage
{\includegraphics[width=0.74\textwidth]{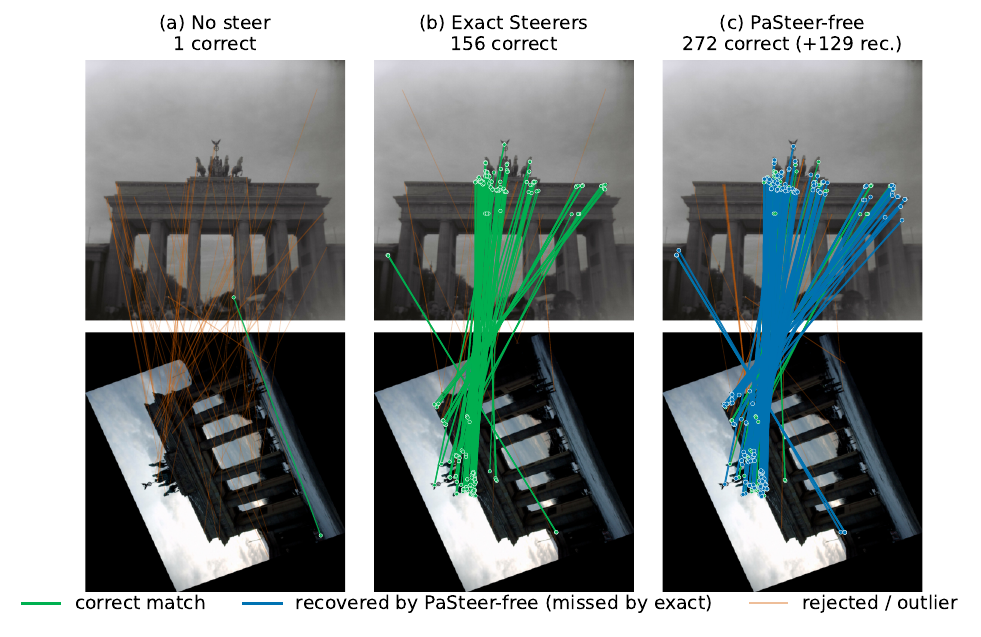}}{fig:qual-md}{%
  \textbf{Qualitative results on MegaDepth-1500~\cite{li2018megadepth}} (scene
  \texttt{0022\_0.5\_0.7}, pair 2), target rotated $112^\circ$ in-plane. No steer: $1$
  correct match; exact Steerers~\cite{bokman2024steerers}: $156$; PaSteer-free: $272$
  ($+129$ recovered).}
{\includegraphics[width=0.74\textwidth]{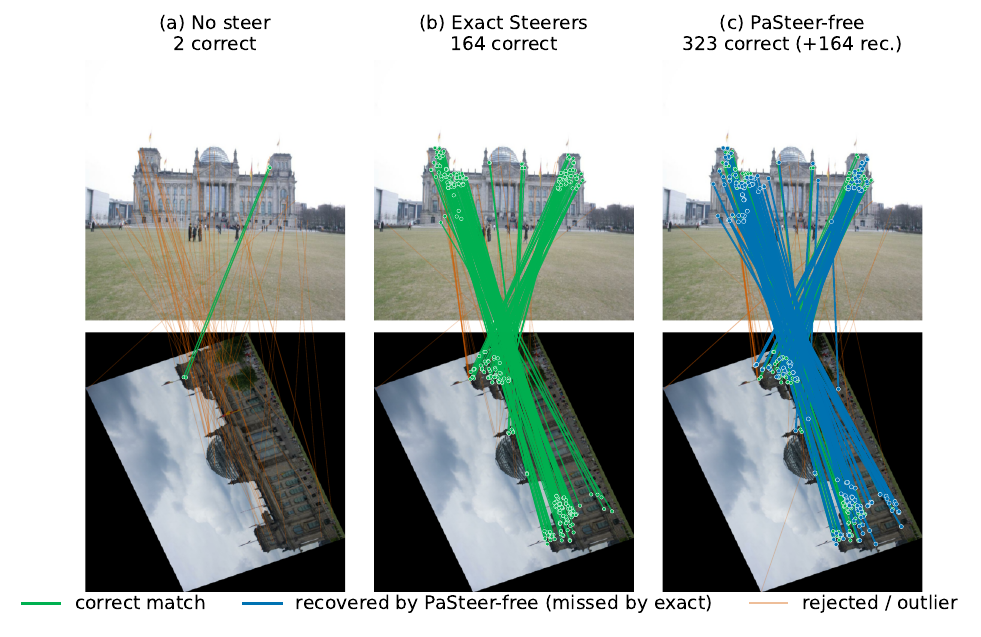}}{fig:qual-imc}{%
  \textbf{Qualitative results on IMC-PT 2020~\cite{jin2021image}} (\texttt{reichstag}),
  target rotated $112^\circ$. No steer: $2$ correct; exact: $164$; PaSteer-free: $323$
  ($+164$ recovered).}

\twofigpage
{\includegraphics[width=0.74\textwidth]{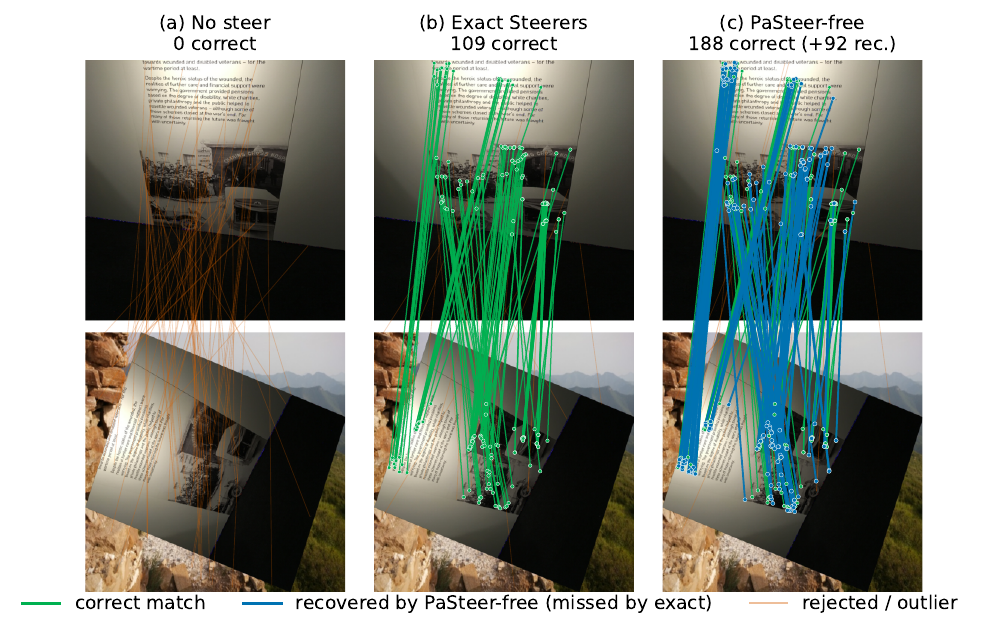}}{fig:qual-roto}{%
  \textbf{Qualitative results on Roto-360~\cite{lee2022rekd}} (sequence \texttt{v\_charing}),
  pure in-plane rotation of $70^\circ$ with a ground-truth homography. No steer: $0$
  correct; exact: $109$; PaSteer-free: $188$ ($+92$ recovered). Inliers are GT-homography
  reprojection $\le 5$\,px.}
{\includegraphics[width=0.74\textwidth]{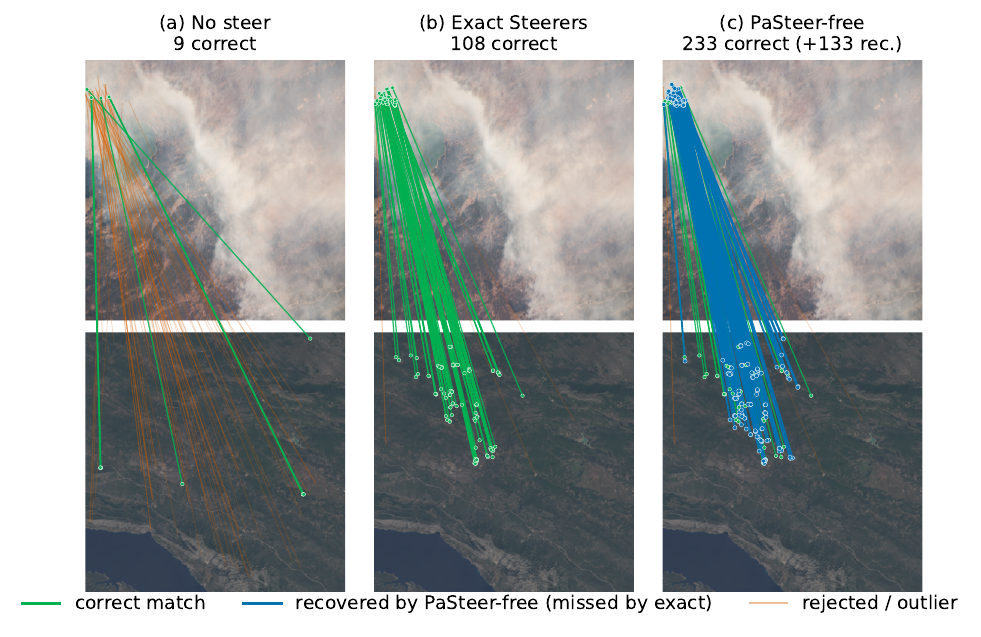}}{fig:qual-aims}{%
  \textbf{Qualitative results on AIMS~\cite{stoken2023aims}} (astronaut frame
  \texttt{ISS052-E-27335} vs.\ a satellite tile), relative rotation $\approx 51^\circ$.
  No steer: $9$ correct; exact: $108$; PaSteer-free: $233$ ($+133$ recovered). Inliers are
  MAGSAC homography inliers.}

\end{document}